\documentclass[10pt,twocolumn,letterpaper]{article}

\usepackage{wacv}
\usepackage{times}
\usepackage{epsfig}
\usepackage{graphicx}
\usepackage{amsmath}
\usepackage{amssymb}
\usepackage{rotating}
\usepackage{array}
\usepackage{setspace}
\usepackage{subfigure}
\usepackage{multicol}
\usepackage{multirow}
\usepackage{adjustbox}
\usepackage{booktabs}
\usepackage{xcolor,colortbl}
\usepackage{color}
\usepackage{makecell}
\usepackage{tabularx}
\usepackage{enumitem}
\usepackage{titling}
\date{\vspace{-3ex}}

\usepackage[accsupp]{axessibility}  

\setlist[itemize]{align=parleft,left=0pt..1em}
\let\tablehead\thead
\renewcommand{\thead}[1]{\textbf{\tablehead{#1}}}

\def\wacvPaperID{0391} 

\wacvfinalcopy 

\ifwacvfinal
\fi

\usepackage[pagebackref=false,breaklinks=true,colorlinks,bookmarks=false]{hyperref}

\begin{document}

\title{Measuring Hidden Bias within Face Recognition via Racial Phenotypes}

\author{Seyma Yucer\textsuperscript{1}, Furkan Tektas\textsuperscript{3}, Noura Al Moubayed\textsuperscript{1} and Toby P. Breckon\textsuperscript{1,2}\\
Department of \{Computer Science\textsuperscript{1}, Engineering\textsuperscript{2}\}, Durham University, Durham, UK\\
BuboAI\textsuperscript{3}, Middlesbrough, UK\\}
\maketitle

\ifwacvfinal
\thispagestyle{empty}
\fi

\begin{abstract}

Recent work reports disparate performance for intersectional racial groups across face recognition tasks: face verification and identification. However, the definition of those racial groups has a significant impact on the underlying findings of such racial bias analysis. Previous studies define these groups based on either demographic information (e.g. African, Asian etc.) or skin tone (e.g. lighter or darker skins). The use of such sensitive or broad group definitions has disadvantages for bias investigation and subsequent counter-bias solutions design. By contrast, this study introduces an alternative racial bias analysis methodology via facial phenotype attributes for face recognition. We use the set of observable characteristics of an individual face where a race-related facial phenotype is hence specific to the human face and correlated to the racial profile of the subject. We propose categorical test cases to investigate the individual influence of those attributes on bias within face recognition tasks. We compare our phenotype-based grouping methodology with previous grouping strategies and show that phenotype-based groupings uncover hidden bias without reliance upon any potentially protected attributes or ill-defined grouping strategies. Furthermore, we contribute corresponding phenotype attribute category labels for two face recognition tasks: RFW for face verification and VGGFace2 (test set) for face identification.

\end{abstract}
\vspace{-0.3cm}
\section{Introduction\label{sec:intro}}
\vspace{-0.1cm}

An increasing number of automated face recognition systems have been deployed by companies, nonprofits and governments to make autonomous decisions for millions of users \cite{kortli2020face}. Such wide-scale adoption within real-world scenarios brings with it valid concerns on the potential abuse of face recognition due to the presence of data and algorithmic bias \cite{garcia2019harms,srinivas2019face}. The most common issue pertaining to such bias arises in racial groups \cite{grother2019face}. Subsequently, the research community have been focused on methods that rely on demographic or skin type group annotations drawn from public face recognition benchmark datasets \cite{cao2018vggface2,guo2016ms}. This provides algorithmic performance on such predefined groupings to measure bias. However, current grouping annotations and related bias evaluation strategies may lead to unintended negative implications- each of which we now detail to illustrate our motivation clearly.

\textbf{Ambiguous Definition of Race:} The historical and biological definitions of race vary and racial context is not fixed over time \cite{lee1994navigating}. Such ambiguity becomes more problematic for the face recognition literature, as many researchers do not provide any related background about the details of their racial categorisation design process \cite{scheuerman2020we}.
However, racial groupings are critical to the effective evolution of face recognition methodologies as they often represent the all-important means of quantitative evaluation. As in any recognition task, poorly defined groupings result in skewed mean and standard deviation measures of relative performance due to the ill-posed boundary conditions on membership of each group that can cause a given an example to justifiably transit from one group to another. 


\textbf{Privacy of Protected Attributes:} Auditing benchmark datasets can cause potential privacy and consent violations \cite{raji2020saving} for dataset subjects. For example, exposing demographic origin may enhance the representations of a group under threat, leading to the potential for racial profiling and associated targeting \cite{mozur2019one}. As information of racial or ethnic origin is sensitive \cite{hepple2010new}, researchers should either avoid revealing such sensitive data or provide an appropriate context for use \cite{raji2020saving}. 

\textbf{Confined Groupings:} Skin or racial grouping strategies such as binary \textit{\{light vs. dark; black vs. white\}} for evaluating racial bias limits the scope of any study as they fail to capture the whole aspect of the bias problem where it needs to consider both multi-racial or less stereotypical members of such groups instead \cite{buolamwini2018gender,mitchell2020diversity} use Fitzpatrick skin type groupings to evaluate racial bias, but one such skin-tone based racial grouping contains multidimensional traits including nose, hair type, eye, and
lips \cite{roth2016multiple}. Leveraging all such traits together instead brings improved interpretations and derivations to address racial bias. 

\textbf{Racial Appearance Bias:} Maddox \cite{maddox2004perspectives} explains racial appearance bias as a negative disposition toward phenotypic variations in facial appearance. He also \cite{maddox2018racial} discusses how race-conscious social policies may fail to address racial bi-

\begin{table*}[!htp]
\begin{tabular}{@{}cccccc@{}}
\toprule\textbf{Study}                & \textbf{\begin{tabular}[c]{@{}c@{}}Dataset \\ Name\end{tabular}} & \textbf{\begin{tabular}[c]{@{}c@{}}Release \\ Year\end{tabular}} & \textbf{Racial Grouping Strategy}                                                                                         & \textbf{\begin{tabular}[c]{@{}c@{}}Number of \\ Image\end{tabular}} & \textbf{Source}                                                                                                                                                                \\ \midrule
\cite{ricanek2006morph}       & MORPH                                                           & 2006                                                    & \begin{tabular}[c]{@{}c@{}}Caucasian, Hispanic, Asian, \\ or African American\end{tabular}                                & 55K                                                   & Public Data                                                                                    \\
\cite{zhifei2017cvpr}         & UTK Face                                                        & 2017                                                    & \begin{tabular}[c]{@{}c@{}}Asian, Black, Indian, Others (like Hispanic, \\ Latino, Middle Eastern) and White\end{tabular} & 20K                                                   & \begin{tabular}[c]{@{}c@{}}MORPH, CACD\\ Web\end{tabular}                                      \\
\cite{wang2019racial}         & RFW                                                             & 2019                                                    & African,Asian, Caucasian,Indian                                                                                           & 45K                                                   & MS-Celeb \cite{guo2016ms}                                                                         \\
\cite{wang2020mitigating}     & BUPT-Balanced                                                   & 2020                                                    & African,Asian, Caucasian,Indian                                                                                           & 1.3M                                                  & MS-Celeb \cite{guo2016ms}                                                                                 \\
\cite{sixta2020fairface}      & \begin{tabular}[c]{@{}c@{}}Fair Face \\ Challenge\end{tabular}  & 2020                                                    & \begin{tabular}[c]{@{}c@{}}Light skin-toned (Fitzpatrick I-III), \\ Dark skin-toned (Fitzpatrick IV-VI)\end{tabular}      & 152K                                                  & \begin{tabular}[c]{@{}c@{}}IJB-C \cite{maze2018iarpa} and \\ public domain images\end{tabular} \\
\cite{hazirbas2021towards}    & \begin{tabular}[c]{@{}c@{}}Casual \\ Conversations\end{tabular} & 2021                                                    & Fitzpatrick Skin Types & 45K*                                                  & Vendor data                                                                                    \\
\cite{karkkainen2019fairface} & FairFace                                                        & 2021                                                    & \begin{tabular}[c]{@{}c@{}}Black, East Asian, Indian, Latino, \\ Middle Eastern, Southeast Asian, and White\end{tabular}  & 108K                                                  & \begin{tabular}[c]{@{}c@{}}Flickr, Twitter\\ Newspapers, Web\end{tabular}                     \\ \hline 
\textbf{Ours}                 & \textbf{VGGFace2 \cite{cao2018vggface2}}                        & \textbf{2018 (2021)}                                           & \textbf{\begin{tabular}[c]{@{}c@{}}Fitzpatrick Skin Types, Nose Shape,\\ Eyelid Type, Lip Shape, Hair Type\end{tabular}}     & \textbf{3.3M}                                         & \textbf{Google Image Search}                                                                   \\
\textbf{Ours}                 & \textbf{RFW \cite{wang2019racial}}                              & \textbf{2019 (2021)}                                           & \textbf{\begin{tabular}[c]{@{}c@{}}Fitzpatrick Skin Types, Nose Shape,\\ Eyelid Type, Lip Shape, Hair Type\end{tabular}}     & \textbf{45K}                                          & \textbf{MS-Celeb \cite{guo2016ms}}                                                                              \\ \bottomrule
\end{tabular}
\vspace{0.2cm}
\caption{Publicly available face datasets for different types of facial analysis tasks and their grouping strategies to address racial bias. \text{*}Casual Conversations dataset provides videos.}
\vspace{-0.6cm}
\label{tab_datasets}
\end{table*}

\noindent
ases in the treatment and outcomes of disadvantaged groups. Many studies show that individuals with more stereotypical racial appearance suffer poorer outcomes than those with less stereotypical appearance for their race \cite{maddox2018racial,skinner2015looking,kahn2011differentially}. On the other hand, a better understanding of the role of phenotypic variation complements solutions for both racial bias \cite{maddox2004perspectives}. By way of phenotype, we mean the set of observable characteristics of an individual face where a race-related facial phenotype is hence specific to the human face and correlated to the racial profile of the subject. 

Accordingly, we propose using race-related facial (phenotype) characteristics within face recognition to investigate racial bias. We categorise representative racial characteristics on the face and explore the impact of each characteristic phenotype attribute: skin types, eyelid type, nose shape, lips shape, hair colour and hair type. We audit these attributes for two different publicly available face datasets: VGGFace2 (test set) and RFW. We assess the impact of both attribute-based and subgroup-based evaluations on racial bias of face recognition tasks. We utilise two different training setups for face verification to compare performance disparities between imbalance and racially balanced training datasets. We compare our phenotype-based evaluation strategy with race or skin type based grouping evaluation. We show that our strategy provides a more elaborate perception of bias without revealing any potentially protected or ill-defined information.

\noindent 
This study presents a new evaluation strategy using facial phenotype attributes to investigate and measure racial bias with greater granularity within face recognition tasks. In this paper, our key contributions are as follows:

\begin{itemize}
\vspace{-0.2cm}

\item we propose a new evaluation strategy that uses facial phenotype attributes rather than race labels to measure racial bias within both attribute-based and subgroup-based performance of state-of-the-art face recognition algorithms.

\vspace{-0.2cm}

\item we contribute additional facial phenotype attribute labelling for the VGGFace2 (face identification) and RFW (face verification) benchmark face datasets.

\vspace{-0.2cm}
\item we uncover the potentially hidden source of bias within the evaluation of racial groups, which is supported by quantitative evidence.
\end{itemize}

\vspace{-0.3cm}
\section{Related Work}
\vspace{-0.1cm}
Automated facial recognition encompasses two different tasks: face identification and face verification. For both tasks, studies present approaches that achieve overall performance on public benchmark datasets \cite{maze2018iarpa, LFWTech} whilst the racial diversity within these datasets is often limited, biased and overlooked \cite{zhou2015naive}. Consequently, numerous studies audit publicly available face datasets to demonstrate the dataset bias in face recognition \cite{wang2019racial, karkkainenfairface, merler2019diversity}. However, the group definitions in use vary, meaning that a lack of consensus makes it significantly harder to tackle bias collaboratively due to an inconsistent problem definition across the literature. 

We show leading benchmark face datasets with their grouping strategies in Table \ref{tab_datasets} to tackle racial bias. As a subset of MS-Celeb-1M, the RFW dataset \cite{wang2019racial} measures the racial performance of face verification on four different racial groups: \textit{\{African, Asian, Indian, Caucasian\}}. FairFace \cite{karkkainenfairface} is another dataset drawn from the YFCC-100M Flickr dataset, providing additional group labels, \textit{\{Middle East, Latino\}} to evaluate bias on wider groupings. UTKFace \cite{zhang2017age} is a large-scale face dataset with five different ethnicity categories for a variety of tasks, such as face detection, age estimation, age progression/regression, etc. More recently, The Casual Conversations Dataset \cite{hazirbas2021towards} yielded from vendor data contains 45K videos with corresponding Fitzpatrick skin type labels. We subsequently categorise studies in the literature according to grouping strategies they adopt and explain each category below. 

\textbf{Racial Groupings:} Although the definition of race carries a large amount of complexity and ambiguity, an increasing number of studies adopt various racial groupings and show performance disparities among them \cite{grother2019face,robinson2020face}. The underlying reasons for such disparate results are summarised into two categories by \cite{cavazos2020accuracy}. Both the distribution of data on pre-defined racial groups and how we measure the bias play a major role in the results \cite{cavazos2020accuracy}. However, the majority of racial bias research rarely contains underlying details about how racial groups are determined or how racial bias evaluation metrics are designed \cite{scheuerman2020we}. Furthermore, \cite{terhorst2021comprehensive} showed that non-explicit racial factors (accessories, hairstyles or facial anomalies) conflates with explicit racial factors (skin tone, nose shape or eye shape) and both factors strongly affect the recognition performance. He discusses the need to investigate each factor in order to have robust, fair, and explainable face recognition solutions. Such needs contradict the use of racial groups as they remain too narrow to have elaborate explanations \cite{barocas2021designing}.

\textbf{Skin Type Groupings}: Moreover, various studies \cite{buolamwini2018gender,sixta2020fairface,hazirbas2021towards} measure the racial bias in face recognition using either the Fitzpatrick Skin Types \cite{fitzpatrick1988validity} or binary skin groups instead of using racial groups. Skin type grouping labels are yielded mostly from crowd-sourcing \cite{merler2019diversity}, neural network-based classifiers \cite{terhorst2020maad,krishnapriya2021analysis} or professional human annotators. Merler \cite{merler2019diversity} presents additional human-interpretable quantitative measures of intrinsic facial features along with subjective annotations. Although \cite{howard2021reliability} show that Fitzpatrick Skin Types classification from uncontrolled imagery is a challenging task, \cite{krishnapriya2021analysis} achieves high correlation with ground-truth labels using consistent reference points in human-level annotation interface. 

Another issue is the impact of skin tone on racial bias. Cook \cite{cook2019demographic} 
shows that the measure of skin reflectance on binary skin groups had the greatest net effect on the average biometric performance of face recognition. On the contrary, \cite{krishnapriya2020issues} claims it is not the case when it comes to continuous Fitzpatrick Skin Type groupings. Furthermore, skin tone is not enough for analysing racial bias as there is no clear evidence that skin tone is the primary driver for disparate false match rates \cite{krishnapriya2020issues}. Accordingly, many studies \cite{muthukumar2018understanding,muthukumar2019color} suggest looking for other race-related facial attributes including lip, eye, face shape in order to measure racial bias within face recognition.

In order to address these aforementioned issues, we propose a phenotype-based evaluation strategy for racial bias within face recognition. We provide facial phenotype attributes for the public benchmark datasets VGGFace2 and RFW in Table \ref{tab_datasets}. We explain the phenotype-based attribute category selection process followed by our annotation framework in Section \ref{sec:5} and \ref{sec:6}, respectively. We elaborate on our findings by measuring two state-of-the-art algorithms performance using imbalance and racially balanced training sets. Firstly, we analyse bias for each phenotype attribute category (attribute-based). Secondly, we produce different appearance-based joint distributions of face subjects and assess algorithm performance on subject groupings (subject-based). We provide related experiments with details in Section \ref{sec:results}.

\vspace{-0.1cm}
\section{Ethical Considerations}

\textbf{Intent:} This work intends to provide a novel racial bias analysis methodology via facial phenotype attributes for face recognition. The proposed strategy avoids the need for researchers to use potentially protected or ill-defined subject attributes and instead introduces racial phenotype attributes to explore racial bias in face recognition.

\textbf{Denotation of Facial Phenotypes:} We denote race-related phenotype attributes according to the studies of \cite{feliciano2016shades,fakhro2015evolution} to have descriptive naming whilst avoiding causing any unintended offence to individuals.

\textbf{Use of VGGFace2 and RFW:} We conduct our experiments on two different face datasets which are publicly available for research use only. The reader is directed to the original source publication and the associated research organisation for access to these datasets. We make available supplementary facial attribute labels for these datasets in order to facilitate the use of our proposed evaluation strategy by other researchers, with the aim of furthering our stated intent above.

\vspace{-0.1cm}
\section{Racial Phenotypes on Face Images \label{sec:5}}
\vspace{-0.1cm}
\begin{figure*}[htbp]
 
 \centering
 \subfigure{
 \includegraphics[width=0.4\linewidth]{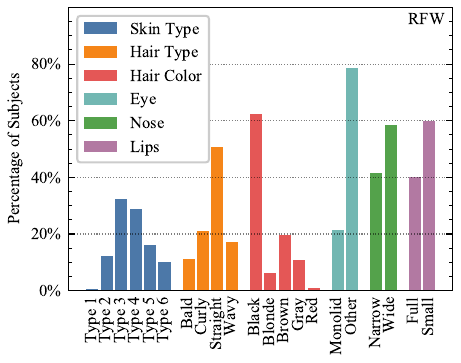}
 }\hspace{1.0cm}
 \subfigure{
 \includegraphics[width=0.4\linewidth]{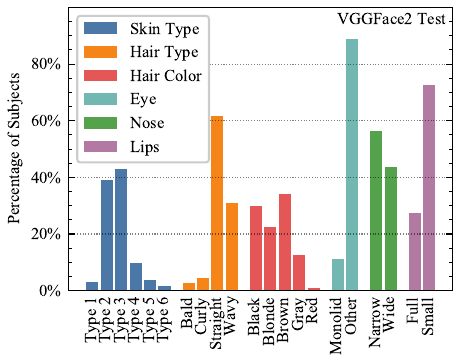}}

 \caption{The distribution of facial phenotype attributes of RFW (left) and VGGFace2 Test (right) datasets.}
 \vspace{-0.6cm}
 \label{fig:dist}
\end{figure*}

In this section, we explain the categorisation of facial phenotype attributes for face recognition. Quine \cite{quine1953three} presents three possible definitions of the race concept: a genetic variation between humans, morphological attributes, and genetically determined psychological characteristics. These morphological attributes are the primary interest for resolving racial bias in face recognition. For morphological attributes, studies \cite{sesardic2010race,ousley2009understanding} focus on the impact of human phenotype characteristics over race estimation. They categorise the attributes by considering biological traits. The study of Shades of Race \cite{feliciano2016shades} investigates the marginal effects of phenotypic characteristics including skin tone, lips, nose, hair and body type on racial categorisation. 
Zhuang \cite{zhuang2010facial} considers 21 anthropometric measurements such as face width, length, nose breadth and length, eye corner points. He finds statistically significant differences in facial measurements between four racial/ethnic groups, which are \textit{\{Caucasian, Hispanic, African, other (mainly Asian)\}}.

We adopt such groupings and measurements for face recognition by considering two limitations. Firstly, effectively evaluating face recognition tasks requires tight cropped (e.g. 112 $\times$ 112 px) low-quality images containing occlusion, shadows, and illumination variations for both the training and test stages. This makes phenotype attribute detection on the specific characteristics of face dataset images more difficult when compared to real-world human faces. Secondly, the broader categorisation increases the number of potential groupings, making bias evaluation inefficient for face recognition systems. Correspondingly, we decide to use 6 primary attributes that define the phenotype groupings for our study: skin type, eyelid type, nose shape, lip shape, hair type and hair colour\footnote{We note that hair information is still present in the tightly cropped images, and it may be helpful for future automated facial analyses tasks.}. Subsequently, we have 21 different attribute categories under the 6 primary attributes as listed in Table \ref{attrtab}.

\begin{table}[ht]

\setlength\tabcolsep{3pt}
\resizebox{\columnwidth}{!}{
\begin{tabular}{l|c|c|c}
\toprule
\textbf{Attribute} & \textbf{Categories} &\textbf{RFW} &\textbf{VGGFace2}\\
\midrule
\textbf{Skin Type} & Type 1 / 2 / 3 / 4 / 5 / 6  & 0.71 & 1.14\\
\textbf{Eyelid Type} & Monolid / Other & 0.80&1.09\\
\textbf{Nose Shape} & Wide / Narrow & 0.24 &0.18 \\
\textbf{Lip Shape} & Full / Small & 0.28 & 0.63 \\
\textbf{Hair Type} & Straight / Wavy / Curly / Bald & 0.70&1.11 \\
\textbf{Hair Colour} & Red / Blonde / Brown / Black / Grey & 1.23 & 0.67 \\ 
\bottomrule
\end{tabular}
}
\vspace{0.1cm}
\caption{Facial phenotype attributes and their categorisation based on \cite{feliciano2016shades} along with normalised standard deviations $\sigma/\mu$.}
\label{attrtab}
\vspace{-0.3cm}
\end{table}

We choose to use Fitzpatrick Skin Types \cite{fitzpatrick1988validity} for skin tones as it provides more granularity, \textit{\{Type 1, Type 2, Type 3, Type 4, Type 5, Type 6\}}, than binary skin-tone groupings, \textit{\{lighter skin-tone, darker skin-tone\}}. The appearance of the human eye has been grouped by its position, shape and settings in many cosmetic industry guidelines \cite{alzahrani2021integrated}. However, they have either no scientific background or solid relation with race. Instead, we look into epicanthal folds and check eyelid difference as it is a more distinctive attribute for racial bias \cite{lee2000anchor}. We acknowledge that a single attribute category can be observed in multiple race groups. However, our main concern is identifying the most observable and convenient racial phenotype attributes on images to evaluate the bias (see Table \ref{attrtab}). 

For the appearance of the nose, we use two categories, wide and narrow, by examining the nasal breadth \cite{zhuang2010facial}. Hair texture is labelled into eight categories using the frequency of twists, waves, and curve diameter metric by \cite{de2007shape}. 
Here we utilise eight categories and group them into three main hair texture types: straight, wavy, curly, in addition to bald. Despite being the most artificially manipulable attribute, we retain hair colour as it is related to skin tone \cite{rees2003genetics}—the categories for hair colour we use: red, grey, black, blonde, brown (see Table \ref{attrtab}).

\vspace{-0.1cm}
\section{Annotation of Racial Phenotypes}

\label{sec:6}
Previously in Section \ref{sec:5}, we explain how we define racial phenotype attributes and their categories. Before the annotation process, we choose the most established face recognition datasets to validate our proposed methodology. For the face verification task, we choose the RFW dataset \cite{wang2019racial} as it provides a relatively broader racial distribution of subjects where each subject contains 3-5 images. For face identification, we use the VGGFace2 closed-test set \cite{cao2018vggface2}, which contains at least 300 images per subject. For both datasets, we design an annotation interface to make the annotation process both user-friendly and robust. We present multiple sample images of a subject to avoid incorrect annotation caused by challenging samples such as grey-scale images, facial makeup, and poor scene illumination. Each subject is presented with attribute category selectors next to a set of face images within the annotation interface.
Subsequently, an experienced annotator who has experience in morphological differences among races annotates each subject using the interface. 

We obtain 11654 subjects annotations from the RFW and VGGFace2 benchmark datasets. Each annotation took 10-20 seconds, and overall annotation took 12 days $($, i.e. annotator working at a maximum of 6 hours per day with regular breaks$)$. The result of this annotation process, the phenotype attributes distributions for the RFW and VGGFace2 benchmark datasets, are shown in Figure \ref{fig:dist} left/right, respectively. We also present the normalised standard deviations (Coefficient of Variance), $\sigma/\mu$, among attribute categories of benchmark datasets to show the level of imbalance within these categories in Table \ref{attrtab}. For both datasets, we can observe that the dominant phenotype attribute categories are Skin Type 3, Straight Hair, Narrow Nose, Other (non-monolid) Eyes, Small Lips, which correlates to the dominant presence of Caucasian faces based on the analysis of Figure \ref{fig:dist}. 
\vspace{-0.1cm}
\section{Experimental Results and Discussion}

\label{sec:results}
In this section, we analyse the performance of our phenotype-based grouping methodology for face recognition tasks. We provide a public reference implementation, dataset reference links and pre-trained models \footnote{\label{note1}\url{https://github.com/seymayucer/FacialPhenotypes}}. 

\subsection{Training Setups}

\noindent
\textbf{Setup 1 (Imbalanced Training Data):}
We train ArcFace \cite{deng2019arcface} with a ResNet100 \cite{he2016deep} on the VGGFace2 benchmark datasets that contains 8631 subjects where subject distribution is racially imbalanced. Here, our specific choice of VGGFace2 is due to investigate the impact of imbalanced training data that includes data bias on our proposed evaluation strategy.

\noindent
\textbf{Setup 2 (Racially Balanced Training Data):} We use a ResNet34 \cite{he2016deep} backbone architecture with the Softmax loss \cite{liu2017sphereface} trained on the BUPT-Balanced benchmark dataset \cite{wang2020mitigating} that contains 28000 face subjects. The BUPT-Balanced has racially balanced distributions among four groups \textit{\{African, Asian, Indian, Caucasian\}} with 7000 face subjects each. The primary purpose of setup 2 is to assess the impact of a racially balanced training dataset on results over the bias using our proposed phenotype-based methodology. We compare how much a racially balanced training dataset improved the performance difference compared to setup 1.

\newcommand*\rot{\rotatebox{90}}

\begin{table}[t]
\small
\centering
\begin{tabularx}{\columnwidth}{Xrr}

\toprule
     \thead{Attribute Name} &   
     \thead{Setup 1 \\ Accuracy  \%} &
     \thead{Setup 2 \\ Accuracy \%} \\
\midrule
  Blonde  Hair &                97.02 &                96.63 \\
     Red  Hair* &                96.33 &                96.83 \\
        Type 2  &                96.22 &                95.83 \\
    Gray  Hair &                94.85 &                95.83 \\
          Bald &                94.75 &                95.70 \\
    Wavy  Hair &                94.32 &                95.50 \\
   Brown  Hair &                94.25 &                94.83 \\
        Type 6  &                93.77 &                94.77 \\
  Narrow  Nose &                92.92 &                94.77 \\

        Type 5  &                92.15 &                94.38 \\
   Curly  Hair &                92.02 &                93.63 \\
   Small  Lips &                91.92 &                94.98 \\
        Type 3 &                91.72 &                93.77 \\
        Type 1* &                91.31 &                89.51 \\
Straight  Hair &                91.25 &                94.32 \\
    Wide  Nose &                90.68 &                91.02 \\

     Full  Lips &                89.98 &                93.23 \\
        Type 4  &                89.90 &                93.55 \\
    Other  Eye &                89.88 &                93.75 \\
   Black  Hair &                89.88 &                91.42 \\
  Monolid  Eye &                88.27 &                89.73 \\
\midrule
     $\sigma$     &  2.44  &   2.06 \\
     $\sigma^{\ast}$ & 2.39     &   1.77\\
    
\bottomrule

\end{tabularx}
\vspace{.1cm}
\caption{Attribute-based face verification performance of RFW. $\sigma$ represents the standard deviation of all attribute category accuracies, including red hair and type 1, $\sigma^{\ast}$ represents excluding standard deviation.}
\label{tab:fr1}
\vspace{-0.6cm}
\end{table}


\begin{table*}[t!]

\centering

\vspace{0.2cm}

\setlength\tabcolsep{4pt} 
\begin{tabular}{cccccrr|cccccrr}
\toprule

\rot{\textbf{Skin}} & \rot{\textbf{Lips}} & \rot{\textbf{Eye}} & \rot{\textbf{Nose}} & \rot{\textbf{Hair Type}} &\multicolumn{1}{c}{\rot{\begin{minipage}{0.5in}\textbf{Ratio}\newline \centering $(\textbf{\%})$ \end{minipage}}} &\multicolumn{1}{c|}{\rot{\begin{minipage}{0.5in}\textbf{Accuracy}\newline \centering $(\textbf{\%})$ \end{minipage}}} & \rot{\textbf{Skin}} & \rot{\textbf{Lips}} & \rot{\textbf{Eye}} & \rot{\textbf{Nose}} & \rot{\textbf{Hair Type}} & \multicolumn{1}{c}{\rot{\begin{minipage}{0.5in}\textbf{Ratio}\newline \centering $(\textbf{\%})$ \end{minipage}}} & \multicolumn{1}{c}{\rot{\begin{minipage}{0.5in}\textbf{Accuracy}\newline \centering $(\textbf{\%})$ \end{minipage}}} \\
\midrule
 $\{1,2\}$ & Small & Other & Narrow & Straight & 3.82 & 96.53 & $\{3,4\}$ & Full & Monolid & Wide & Straight & 1.55 & 91.63 \\
 $\{3,4\}$ & Small & Other & Narrow & Straight & 7.43 & 96.45 & $\{1,2\}$ & Small & Other & Narrow & Bald & 0.28 & 91.29 \\
 $\{3,4\}$ & Small & Other & Narrow & Wavy & 3.67 & 96.11 & $\{5,6\}$ & Full & Other & Narrow & Curly & 1.97 & 91.23 \\
 $\{1,2\}$ & Small & Other & Wide & Straight & 3.03 & 95.63 & $\{3,4\}$ & Small & Other & Wide & Bald & 1.68 & 91.01 \\
 $\{1,2\}$ & Small & Other & Narrow & Wavy & 1.64 & 95.62 & $\{1,2\}$ & Full & Other & Narrow & Wavy & 0.27 & 90.74 \\
 $\{1,2\}$ & Full & Other & Narrow & Straight & 0.70 & 95.59 & $\{3,4\}$ & Small & Monolid & Wide & Wavy & 0.96 & 90.17 \\
 $\{3,4\}$ & Full & Other & Narrow & Straight & 3.59 & 95.28 & $\{1,2\}$ & Small & Other & Wide & Bald & 0.46 & 89.78 \\
 $\{3,4\}$ & Full & Other & Wide & Straight & 4.47 & 94.98 & $\{5,6\}$ & Small & Other & Narrow & Curly & 0.81 & 89.50 \\
 $\{3,4\}$ & Small & Other & Wide & Wavy & 2.95 & 94.92 & $\{3,4\}$ & Small & Monolid & Narrow & Wavy & 1.20 & 89.35 \\
 $\{3,4\}$ & Small & Other & Wide & Straight & 8.83 & 94.92 & $\{5,6\}$ & Full & Other & Wide & Curly & 13.09 & 89.18 \\
 $\{1,2\}$ & Full & Other & Wide & Straight & 0.33 & 94.87 & $\{3,4\}$ & Full & Other & Wide & Bald & 0.80 & 86.02 \\
 $\{1,2\}$ & Small & Other & Wide & Wavy & 0.72 & 94.56 & $\{5,6\}$ & Small & Other & Wide & Bald & 0.99 & 85.90 \\
 $\{3,4\}$ & Small & Other & Wide & Curly & 0.51 & 93.89 & $\{3,4\}$ & Full & Other & Wide & Curly & 0.46 & 85.38 \\
 $\{3,4\}$ & Full & Other & Wide & Wavy & 1.90 & 93.41 & $\{3,4\}$ & Small & Monolid & Narrow & Bald & 0.32 & 84.10 \\
 $\{3,4\}$ & Full & Other & Narrrow & Wavy & 1.94 & 93.10 & $\{5,6\}$ & Small & Other & Narrow & Bald & 0.30 & 82.81 \\
 $\{3,4\}$ & Small & Other & Narrow & Bald & 0.68 & 92.50 & $\{3,4\}$ & Small & Monolid & Wide & Bald & 0.52 & 82.67 \\
 $\{3,4\}$ & Small & Other & Narrow & Curly & 0.31 & 92.45 & $\{3,4\}$ & Full & Monolid & Narrow & Wavy & 0.43 & 82.04 \\
 $\{5,6\}$ & Small & Other & Wide & Curly & 2.81 & 92.23 & $\{5,6\}$ & Full & Other & Narrow & Bald & 0.53 & 81.24 \\
 $\{3,4\}$ & Small & Monolid & Wide & Straight & 6.59 & 91.93 & $\{1,2\}$ & Small & Monolid & Narrow & Straight & 0.47 & 81.04 \\
 $\{3,4\}$ & Full & Monolid & Narrow & Straight & 1.81 & 91.78 & $\{3,4\}$ & Full & Monolid & Wide & Wavy & 0.27 & 79.47 \\
 $\{5,6\}$ & Full & Other & Wide & Bald & 3.62 & 91.74 & $\{5,6\}$ & Full & Other & Wide & Wavy & 0.32 & 78.94 \\
 $\{3,4\}$ & Small & Monolid & Narrow & Straight & 7.95 & 91.70 & & & & & & & \\
\midrule
\multicolumn{13}{l}{$\sigma$}   &   5.07 \\
\bottomrule 
\end{tabular}
\vspace{0.2cm}
\caption{Subgroup-based face verification performance of RFW using training setup 1, sorted by descending order of accuracy.\label{tab:4} }
\vspace{-0.6cm}
\end{table*}

\subsection{Face Verification}
\label{sec:fver}
Face verification, also known as one-to-one verification, is the task of comparing two different facial images to estimate whether they belong to the same individual subject. We follow two pairing strategies to explore the impact of a single attribute (attribute-based) and appearance-based facial groups (subgroup-based) on the evaluation performance of face verification.

\noindent
\textbf{Attribute-based pairing:} Firstly, we generate pairs from images containing the same attribute category—for example, facial images from people who all have monolid eyes. Consequently, we compare individual attributes performance using both training setups for face verification.

For attribute-based face verification, we randomly select 20k positive and 20k negative pairs from all possible pairs of each attribute. We calculate the cosine similarity of feature encoding of all selected negative and positive pairs to obtain the most challenging pairs. Subsequently, we select the most similar 3000 pairs from the negative samples and the least similar 3000 pairs from the positive samples for each attribute category in Table \ref{tab:fr1}. Since the Type 1 category of the skin type attribute and red hair category of hair colour attribute do not have enough samples to generate 6000 pairs, we instead produce 602 pairs (301 positive, 301 negative) for Type 1, 1200 (600 positives, 600 negative) pairs for red hair.

In this way, we measure each face attributes accuracy using on face verification performance. We use both training setups to show how much standard deviation (\,$\sigma$)\ changes between balanced and imbalanced training data. We present the attribute-based sample groups in Table \ref{tab:fr1} with a standard deviation of accuracies excluding red hair and Type 1 attribute accuracies (\,$\sigma*$)\, and including them (\,$\sigma$)\,. It is clear from Table \ref{tab:fr1} that for both setup 1 (imbalanced training data) and setup 2 (racially balanced training data), accuracy is lower for monolid eyes, black hair, full lips, and wide nose than the other eye, blonde hair, and small lips, and narrow nose respectively. We also do find a slight correlation between darker skin tones and higher false matching rates when we pair from the same attribute categories (Supplementary Table \ref{tab:tnr}). Moreover, although the imbalanced training setup results a bigger performance difference (\,$\sigma$)\,, the amount of difference between two setups is small, meaning that a racially balanced dataset distribution is not enough to overcome performance bias.

\begin{figure*}[!t]
  
  \centering
  
  \includegraphics[width=\linewidth]{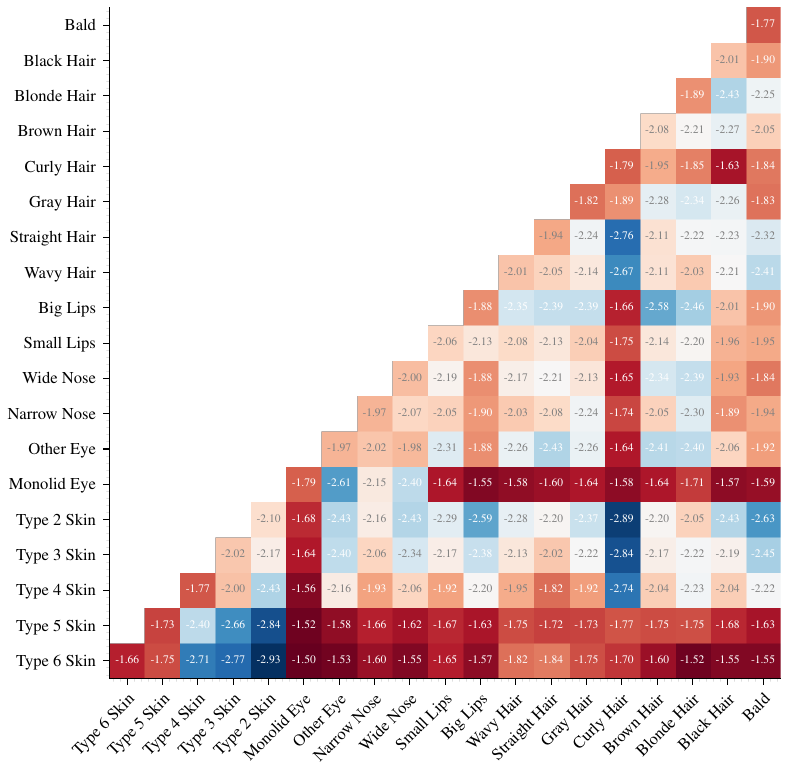}
  
  \caption{False matching rates (FMR) of cross-attribute based pairings for 21 attribute categories using training setup 1. Each cell depicts FMR on a logarithmic scale which is log10(FMR) with lower negative values (close to zero) encoding superior false match rates.}
  
  \label{fig:covariance}
  \vspace{-0.5cm}
\end{figure*}

Additionally, NIST \cite{grother2019face} suggests providing false matching rates of pairing combinations between each grouping in the dataset as it is necessary for real-world scenarios. Therefore, we pair each attribute category with all other attribute categories to assess cross-attribute pairing performance. Subsequently, we evaluate false matching rates between any attribute category pair combination in Figure \ref{fig:covariance}. We randomly generate 10000 pairs for each category pairings; in total, we have 441 (21 $\times$ 21) pairings. For example, each cross-attribute pairings means 10000 pairs between blonde hair - monolid eye, type 3 - wide nose or wavy hair - full lips etc.
As a result of this, we clearly show that Type 5, Type 6 and monolid eyes pairings have higher false matching rates among all attribute categories in Figure \ref{fig:covariance} using training setup 1. Consequently, the impact of the dark skin tones on performance increases for cross-attribute pairings compared to the attribute-based pairings.

\noindent
\textbf{Subgroup-based pairing:} Secondly, we create various subgroups with different phenotypic attribute combinations in the dataset. For example, one such subgroup consists of subjects with skin type 3, monolid eyes, straight hair, wide nose, and small lips. Our main purpose of such pairing is to show the effects of single attribute changes over a group-for instance, what would change when only skin gets darker, but other attributes remain the same?

Furthermore, we generate all possible subgroups with different phenotypic attribute category combinations to investigate subgroup-based performances. However, we need to limit the number of subgroups such that we can present our results efficiently. We first remove the hair colour attribute as it is the easiest race-relevant attribute that individuals can readily modify via styling. Consequently, we merge skin types into three groups and show them as \{1,2\} for Type 1 and Type 2, \{3,4\} for Type 3 and Type 4, and \{5,6\} for Type 5 and Type 6. Lastly, we remove subgroups with a few or even no samples in the test set, which comprises 3\% of all samples. In Table \ref{tab:4}, we show the performance of each subgroup with its proportion in the original test dataset. To evaluate the performance, we generate 6000 pairs (3k positive and 3k negative) from all possible pairs of subgroups that have enough samples. For the rest, we generate an equal number of negative and positive pairs as much as availability facilitates. From our observation of Table \ref{tab:4}, we can conclude that groups who have one of the attributes like wide nose, full lips, and monolid eye type always have less accuracy than the other groups with a narrow nose, small lips and other eye (when rest of the attributes are same). Furthermore, whilst the average accuracy of the subgroups with Type \{5,6\} skin type is 86.97\%, subgroups with Type \{1,2\} skin type is 92.56\%, but this notably includes other attributes effects.

Moreover, the number of subgroup variations with darker skin tones are much smaller than lighter tones which causes many different evaluation and analysis problems. It lacks sufficient interpretation in the test phase; there are minorities in the global populous with dark skin and monolid eyes or any other less common variations. Benchmark datasets do not contain enough representations for such minority groups. An improved evaluation dataset would be one that is able to cover more phenotype combinations such that its distribution is an unbiased representation of the global populous.

\begin{figure}[!t]
  
  \centering
  
  \includegraphics[width=\linewidth]{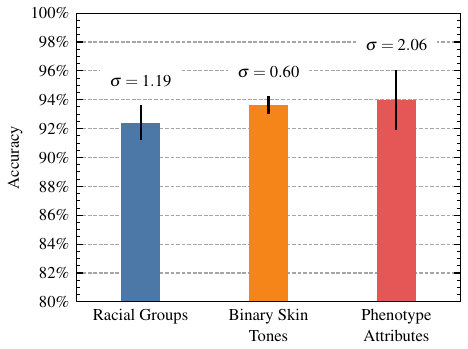}
  
  \caption{Accuracy variations for three grouping strategies. Standard deviation of the groupings reflects the amount of measured bias. Racial groupings \textit{\{African, Asian, Caucasian, Indian\}} accuracies are obtained from \cite{wang2020mitigating}. Binary skin tones \textit{\{lighter skin-tone, darker skin-tone\}} are the average accuracy of Type 1-3 and Type 4-6 skin tones, respectively.}
  \label{fig:std}
  \vspace{-0.7cm}
\end{figure}

Lastly, we estimate such disparities among different grouping strategies using training setup 2. We take racial groupings \textit{\{African, Asian, Indian, Caucasian\}} and binary skin tone groupings \textit{\{lighter skin-tone, darker skin-tone\}} as they are very common grouping strategies in the literature. We compare them with our phenotype-based grouping strategy. In Figure \ref{fig:std}, we show that how accuracy and the standard deviation differs between sub-groups in three different strategies. Higher variation reveals hidden bias, which may be missed in narrow, erroneous racial or binary skin tones grouping strategies. The phenotype-based grouping strategy brings a more granular observation of the variability in performance (i.e. higher standard deviation) and hence a more resolute measure of performance bias.

\vspace{-0.1cm}
\subsection{Face Identification}
\vspace{-0.1cm}
Face identification as a one-to-many verification is the task of searching for a face across a facial database. There are two scenarios for face identification applications based on whether a queried face is enrolled in a database or not. Open-set identification assumes the database does not necessarily contain the queried face, while closed-set identification always looks for a match in the database.
In this study, we apply closed-set identification using the test set of the VGGFace2 benchmark dataset on the originally proposed protocol \cite{cao2018vggface2} and we extract the image features using training setup 1 \cite{deng2019arcface}. We apply a 5-fold train-test split where we sample 50 images from each subject as the test set and use the rest as the training set. We train a standard linear SVM on the extracted feature representations and predict the identities for test samples. Our results are shown in Table \ref{tab:faceid} where we can observe that the standard deviation (\,$\sigma$)\, is much smaller when compared to the earlier attribute-based face verification results of Table \ref{tab:fr1}. It shows that the closed-set face identification does not have the same level of bias correlation as we find for face verification. However, in this experiment, we are unable to have the same proportion for each attribute, and we did not measure open-set face identification. As suggested in \cite{grother2019face}, future work should design and apply open-set tests for face identification on better-distributed benchmark datasets to measure bias extensively. 

\begin{table}[t!]
\small
\centering

\setlength\tabcolsep{2pt} 
\resizebox{\columnwidth}{!}{
\begin{tabular}{lrr|lrr}
\toprule
        \textbf{Attribute } &  \textbf{Ratio (\%)} & \textbf{ Acc (\%)} &     \textbf{ Attribute} &  \textbf{ Ratio (\%)} & \textbf{Acc (\%)} \\
\midrule
     Bald  &   2.80 &    97.49 &        Type 6  &   1.60 &    96.25 \\
     Grey Hair  &  12.60 &    97.47 &     Wide Nose &  43.60 &    96.19 \\
      Red Hair  &   0.80 &    97.10 &        Type 3  &  42.80 &    96.13 \\
         Type 5  &   3.60 &    96.87 &   Brown Hair  &  34.20 &    96.05 \\
         Type 4  &   9.60 &    96.75 &   Curly Hair  &   4.40 &    95.93 \\
     Small Lips &  72.60 &    96.56 &    Wavy Hair  &  31.00 &    95.92 \\
         Type 2  &  39.20 &    96.43 &    Monolid Eye &  11.00 &    95.73 \\
    Black Hair  &  29.80 &    96.43 &  Blonde Hair  &  22.60 &    95.52 \\
 Straight Hair  &  61.80 &    96.35 &      Full Lips &  27.40 &    95.36 \\
     Other Eye &  89.00 &    96.29 &        Type 1  &   3.20 &    92.90 \\
    Narrow Nose &  56.40 &    96.26 &               &        &          \\       
\hline
    \multicolumn{5}{l}{$\sigma$} & 0.93 \\
\bottomrule 
\end{tabular}}
\vspace{0.2cm}
\caption{Face identification performance on VGGFace2 test set using standard linear SVM and features from training setup 1, sorted by descending order of accuracy.\label{tab:faceid}}
\vspace{-0.5cm}

\end{table}

\vspace{-0.1cm}
\section{Conclusion}

We propose a new evaluation strategy using facial phenotype attributes to assess racial bias in face recognition tasks. We elaborate experimental results to show the impact of each phenotype attributes using two different training setups, including imbalanced and racially balanced training sets. We also provide different pairing strategies for face verification to draw attention to the importance of pairing for comprehensive evaluation. We observe apparent performance differences between race-related phenotype attribute categories and subgroups for both training setups. However, we also uncover more considerable performance disparities among phenotype attributes than racial groups. We demonstrate that phenotype-based evaluation strategy reveals racial bias comprehensively whilst avoiding exposing potentially protected or ill-defined attributes. Future work will focus on improving facial appearance variations using generative models to provide more balanced and realistic test scenario distributions.

\clearpage 
{\small

\bibliography{egbib}

\begin{thebibliography}{10}

\bibitem{kortli2020face}
Yassin Kortli, Maher Jridi, Ayman Al~Falou, and Mohamed Atri.
\newblock Face recognition systems: A survey.
\newblock {\em Sensors}, 2020.

\bibitem{garcia2019harms}
Raul~Vicente Garcia, Lukasz Wandzik, Louisa Grabner, and Joerg Krueger.
\newblock The harms of demographic bias in deep face recognition research.
\newblock In {\em 2019 International Conference on Biometrics (ICB)}. IEEE,
  2019.

\bibitem{srinivas2019face}
Nisha Srinivas, Karl Ricanek, Dana Michalski, David~S Bolme, and Michael King.
\newblock Face recognition algorithm bias: Performance differences on images of
  children and adults.
\newblock In {\em Proceedings of the IEEE/CVF Conference on Computer Vision and
  Pattern Recognition Workshops}, 2019.

\bibitem{grother2019face}
Patrick Grother, Mei Ngan, and Kayee Hanaoka.
\newblock {\em Face Recognition Vendor Test (FVRT): Part 3, Demographic
  Effects}.
\newblock National Institute of Standards and Technology, 2019.

\bibitem{cao2018vggface2}
Qiong Cao, Li~Shen, Weidi Xie, Omkar~M Parkhi, and Andrew Zisserman.
\newblock Vggface2: A dataset for recognising faces across pose and age.
\newblock In {\em IEEE International Conference on Automatic Face \& Gesture
  Recognition}, 2018.

\bibitem{guo2016ms}
Yandong Guo, Lei Zhang, Yuxiao Hu, Xiaodong He, and Jianfeng Gao.
\newblock Ms-celeb-1m: A dataset and benchmark for large-scale face
  recognition.
\newblock In {\em European conference on computer vision}. Springer, 2016.

\bibitem{lee1994navigating}
Jayne Chong-Soon Lee.
\newblock Navigating the topology of race, 1994.

\bibitem{scheuerman2020we}
Morgan~Klaus Scheuerman, Kandrea Wade, Caitlin Lustig, and Jed~R Brubaker.
\newblock How we've taught algorithms to see identity: Constructing race and
  gender in image databases for facial analysis.
\newblock {\em ACM on Human-Computer Interaction}, 2020.

\bibitem{raji2020saving}
Inioluwa~Deborah Raji, Timnit Gebru, Margaret Mitchell, Joy Buolamwini,
  Joonseok Lee, and Emily Denton.
\newblock Saving face: Investigating the ethical concerns of facial recognition
  auditing.
\newblock In {\em AAAI/ACM Conference on AI, Ethics, and Society}, 2020.

\bibitem{mozur2019one}
Paul Mozur.
\newblock One month, 500,000 face scans: How china is using ai to profile a
  minority.
\newblock {\em The New York Times}, 2019.

\bibitem{hepple2010new}
Bob Hepple.
\newblock The new single equality act in britain.
\newblock {\em The Equal Rights Review}, 2010.

\bibitem{buolamwini2018gender}
Joy Buolamwini and Timnit Gebru.
\newblock Gender shades: Intersectional accuracy disparities in commercial
  gender classification.
\newblock In {\em Conference on Fairness, Accountability and Transparency},
  2018.

\bibitem{mitchell2020diversity}
Margaret Mitchell, Dylan Baker, Nyalleng Moorosi, Emily Denton, Ben Hutchinson,
  Alex Hanna, Timnit Gebru, and Jamie Morgenstern.
\newblock Diversity and inclusion metrics in subset selection.
\newblock In {\em AAAI/ACM Conference on AI, Ethics, and Society}, 2020.

\bibitem{roth2016multiple}
Wendy~D Roth.
\newblock The multiple dimensions of race.
\newblock {\em Ethnic and Racial Studies}, 2016.

\bibitem{maddox2004perspectives}
Keith~B Maddox.
\newblock Perspectives on racial phenotypicality bias.
\newblock {\em Personality and Social Psychology Review}, 2004.

\bibitem{maddox2018racial}
Keith~B Maddox and Jennifer~M Perry.
\newblock Racial appearance bias: Improving evidence-based policies to address
  racial disparities.
\newblock {\em Policy Insights from the Behavioral and Brain Sciences}, 2018.

\bibitem{ricanek2006morph}
Karl Ricanek and Tamirat Tesafaye.
\newblock Morph: A longitudinal image database of normal adult age-progression.
\newblock In {\em 7th International Conference on Automatic Face and Gesture
  Recognition (FGR06)}. IEEE, 2006.

\bibitem{zhifei2017cvpr}
Zhifei Zhang, Yang Song, and H.~Qi.
\newblock Age progression/regression by conditional adversarial autoencoder.
\newblock {\em 2017 IEEE Conference on Computer Vision and Pattern Recognition
  (CVPR)}, 2017.

\bibitem{wang2019racial}
Mei Wang, Weihong Deng, Jiani Hu, Xunqiang Tao, and Yaohai Huang.
\newblock Racial faces in the wild: Reducing racial bias by information
  maximization adaptation network.
\newblock In {\em IEEE International Conference on Computer Vision}, 2019.

\bibitem{wang2020mitigating}
Mei Wang and Weihong Deng.
\newblock Mitigating bias in face recognition using skewness-aware
  reinforcement learning.
\newblock In {\em IEEE Conference on Computer Vision and Pattern Recognition},
  2020.

\bibitem{sixta2020fairface}
Tom{\'a}{\v{s}} Sixta, Julio Junior, CS~Jacques, Pau Buch-Cardona, Eduard
  Vazquez, and Sergio Escalera.
\newblock Fairface challenge at eccv 2020: Analyzing bias in face recognition.
\newblock {\em arXiv preprint arXiv:2009.07838}, 2020.

\bibitem{maze2018iarpa}
Brianna Maze, Jocelyn Adams, James~A Duncan, Nathan Kalka, Tim Miller, Charles
  Otto, Anil~K Jain, W~Tyler Niggel, Janet Anderson, Jordan Cheney, et~al.
\newblock Iarpa janus benchmark-c: Face dataset and protocol.
\newblock In {\em IEEE International Conference on Biometrics}, 2018.

\bibitem{hazirbas2021towards}
Caner Hazirbas, Joanna Bitton, Brian Dolhansky, Jacqueline Pan, Albert Gordo,
  and Cristian~Canton Ferrer.
\newblock Casual conversations: A dataset for measuring fairness in ai.
\newblock In {\em Proceedings of the IEEE/CVF Conference on Computer Vision and
  Pattern Recognition}, 2021.

\bibitem{karkkainen2019fairface}
Kimmo K{\"a}rkk{\"a}inen and Jungseock Joo.
\newblock Fairface: Face attribute dataset for balanced race, gender, and age.
\newblock {\em arXiv preprint arXiv:1908.04913}, 2019.

\bibitem{skinner2015looking}
Allison~L Skinner and Gandalf Nicolas.
\newblock Looking black or looking back? using phenotype and ancestry to make
  racial categorizations.
\newblock {\em Journal of Experimental Social Psychology}, 2015.

\bibitem{kahn2011differentially}
Kimberly~Barsamian Kahn and Paul~G Davies.
\newblock Differentially dangerous? phenotypic racial stereotypicality
  increases implicit bias among ingroup and outgroup members.
\newblock {\em Group Processes \& Intergroup Relations}, 2011.

\bibitem{LFWTech}
Gary~B. Huang, Manu Ramesh, Tamara Berg, and Erik Learned-Miller.
\newblock Labeled faces in the wild: A database for studying face recognition
  in unconstrained environments.
\newblock Technical report, University of Massachusetts, Amherst, 2007.

\bibitem{zhou2015naive}
Erjin Zhou, Zhimin Cao, and Qi~Yin.
\newblock Naive-deep face recognition: Touching the limit of lfw benchmark or
  not?
\newblock {\em arXiv preprint arXiv:1501.04690}, 2015.

\bibitem{karkkainenfairface}
Kimmo Karkkainen and Jungseock Joo.
\newblock Fairface: Face attribute dataset for balanced race, gender, and age
  for bias measurement and mitigation.
\newblock In {\em IEEE/CVF Winter Conference on Applications of Computer
  Vision)}, 2021.

\bibitem{merler2019diversity}
Michele Merler, Nalini Ratha, Rogerio~S Feris, and John~R Smith.
\newblock Diversity in faces.
\newblock {\em arXiv preprint arXiv:1901.10436}, 2019.

\bibitem{zhang2017age}
Zhifei Zhang, Yang Song, and Hairong Qi.
\newblock Age progression/regression by conditional adversarial autoencoder.
\newblock In {\em IEEE Conference on Computer Vision and Pattern Recognition},
  2017.

\bibitem{robinson2020face}
Joseph~P Robinson, Gennady Livitz, Yann Henon, Can Qin, Yun Fu, and Samson
  Timoner.
\newblock Face recognition: too bias, or not too bias?
\newblock In {\em Proceedings of the IEEE/CVF Conference on Computer Vision and
  Pattern Recognition Workshops}, 2020.

\bibitem{cavazos2020accuracy}
Jacqueline~G Cavazos, P~Jonathon Phillips, Carlos~D Castillo, and Alice~J
  O’Toole.
\newblock Accuracy comparison across face recognition algorithms: Where are we
  on measuring race bias?
\newblock {\em IEEE Transactions on Biometrics, Behavior, and Identity
  Science}, 2020.

\bibitem{terhorst2021comprehensive}
Philipp Terh{\"o}rst, Jan~Niklas Kolf, Marco Huber, Florian Kirchbuchner, Naser
  Damer, Aythami Morales, Julian Fierrez, and Arjan Kuijper.
\newblock A comprehensive study on face recognition biases beyond demographics.
\newblock {\em arXiv preprint arXiv:2103.01592}, 2021.

\bibitem{barocas2021designing}
Solon Barocas, Anhong Guo, Ece Kamar, Jacquelyn Krones, Meredith~Ringel Morris,
  Jennifer~Wortman Vaughan, Duncan Wadsworth, and Hanna Wallach.
\newblock Designing disaggregated evaluations of ai systems: Choices,
  considerations, and tradeoffs.
\newblock {\em arXiv preprint arXiv:2103.06076}, 2021.

\bibitem{fitzpatrick1988validity}
Thomas~B Fitzpatrick.
\newblock The validity and practicality of sun-reactive skin types i through
  vi.
\newblock {\em Archives of dermatology}, 1988.

\bibitem{terhorst2020maad}
Philipp Terh{\"o}rst, Daniel F{\"a}hrmann, Jan~Niklas Kolf, Naser Damer,
  Florian Kirchbuchner, and Arjan Kuijper.
\newblock Maad-face: A massively annotated attribute dataset for face images.
\newblock {\em arXiv preprint arXiv:2012.01030}, 2020.

\bibitem{krishnapriya2021analysis}
KS~Krishnapriya, Michael~C King, and Kevin~W Bowyer.
\newblock Analysis of manual and automated skin tone assignments for face
  recognition applications.
\newblock {\em arXiv preprint arXiv:2104.14685}, 2021.

\bibitem{howard2021reliability}
John~J Howard, Yevgeniy~B Sirotin, Jerry~L Tipton, and Arun~R Vemury.
\newblock Reliability and validity of image-based and self-reported skin
  phenotype metrics.
\newblock {\em arXiv preprint arXiv:2106.11240}, 2021.

\bibitem{cook2019demographic}
Cynthia~M Cook, John~J Howard, Yevgeniy~B Sirotin, Jerry~L Tipton, and Arun~R
  Vemury.
\newblock Demographic effects in facial recognition and their dependence on
  image acquisition: An evaluation of eleven commercial systems.
\newblock {\em IEEE Transactions on Biometrics, Behavior, and Identity
  Science}, 2019.

\bibitem{krishnapriya2020issues}
KS~Krishnapriya, V{\'\i}tor Albiero, Kushal Vangara, Michael~C King, and
  Kevin~W Bowyer.
\newblock Issues related to face recognition accuracy varying based on race and
  skin tone.
\newblock {\em IEEE Transactions on Technology and Society}, 2020.

\bibitem{muthukumar2018understanding}
Vidya Muthukumar, Tejaswini Pedapati, Nalini Ratha, Prasanna Sattigeri,
  Chai-Wah Wu, Brian Kingsbury, Abhishek Kumar, Samuel Thomas, Aleksandra
  Mojsilovic, and Kush~R Varshney.
\newblock Understanding unequal gender classification accuracy from face
  images.
\newblock {\em arXiv preprint arXiv:1812.00099}, 2018.

\bibitem{muthukumar2019color}
Vidya Muthukumar.
\newblock Color-theoretic experiments to understand unequal gender
  classification accuracy from face images.
\newblock In {\em Proceedings of the IEEE/CVF Conference on Computer Vision and
  Pattern Recognition Workshops}, 2019.

\bibitem{feliciano2016shades}
Cynthia Feliciano.
\newblock Shades of race: How phenotype and observer characteristics shape
  racial classification.
\newblock {\em American Behavioral Scientist}, 2016.

\bibitem{fakhro2015evolution}
Abdulla Fakhro, Hyung~Woo Yim, Yong~Kyu Kim, and Anh~H Nguyen.
\newblock The evolution of looks and expectations of asian eyelid and eye
  appearance.
\newblock In {\em Seminars in plastic surgery}. Thieme Medical Publishers,
  2015.

\bibitem{quine1953three}
Willard~V Quine.
\newblock Three grades of modal involvment.
\newblock In {\em XIth International Congress of Philosophy}, 1953.

\bibitem{sesardic2010race}
Neven Sesardic.
\newblock Race: a social destruction of a biological concept.
\newblock {\em Biology \& Philosophy}, 2010.

\bibitem{ousley2009understanding}
Stephen Ousley, Richard Jantz, and Donna Freid.
\newblock Understanding race and human variation: why forensic anthropologists
  are good at identifying race.
\newblock {\em American Journal of Physical Anthropology}, 2009.

\bibitem{zhuang2010facial}
Ziqing Zhuang, Douglas Landsittel, Stacey Benson, Raymond Roberge, and Ronald
  Shaffer.
\newblock Facial anthropometric differences among gender, ethnicity, and age
  groups.
\newblock {\em Annals of occupational hygiene}, 2010.

\bibitem{alzahrani2021integrated}
Theiab Alzahrani, Waleed Al-Nuaimy, and Baidaa Al-Bander.
\newblock Integrated multi-model face shape and eye attributes identification
  for hair style and eyelashes recommendation.
\newblock {\em Computation}, 2021.

\bibitem{lee2000anchor}
Yoonho Lee, Euitae Lee, and Won~Jin Park.
\newblock Anchor epicanthoplasty combined with out-fold type double
  eyelidplasty for asians: do we have to make an additional scar to correct the
  asian epicanthal fold?
\newblock {\em Plastic and reconstructive surgery}, 2000.

\bibitem{de2007shape}
Roland De~La~Mettrie, Didier Saint-L{\'e}ger, Geneviev{\`e}ve Loussouarn,
  Annelise Garcel, Crystal Porter, and Andr{\'e} Langaney.
\newblock Shape variability and classification of human hair: a worldwide
  approach.
\newblock {\em Human biology}, 2007.

\bibitem{rees2003genetics}
Jonathan~L Rees.
\newblock Genetics of hair and skin color.
\newblock {\em Annual review of genetics}, 2003.

\bibitem{deng2019arcface}
Jiankang Deng, Jia Guo, Niannan Xue, and Stefanos Zafeiriou.
\newblock Arcface: Additive angular margin loss for deep face recognition.
\newblock In {\em IEEE International Conference on Computer Vision and Pattern
  Recognition}, 2019.

\bibitem{he2016deep}
Kaiming He, Xiangyu Zhang, Shaoqing Ren, and Jian Sun.
\newblock Deep residual learning for image recognition.
\newblock In {\em IEEE Conference on Computer Vision and Pattern Recognition},
  2016.

\bibitem{liu2017sphereface}
Weiyang Liu, Yandong Wen, Zhiding Yu, Ming Li, Bhiksha Raj, and Le~Song.
\newblock Sphereface: Deep hypersphere embedding for face recognition.
\newblock In {\em Proceedings of the IEEE conference on computer vision and
  pattern recognition}, 2017.

\end{thebibliography}
}




\pagenumbering{gobble}
\title{Measuring Hidden Bias within Face Recognition via Racial Phenotypes - Supplementary Material}

\maketitle

\section{Supplementary Material}

\setcounter{table}{0}
\setcounter{figure}{0}
\renewcommand{\thetable}{S\arabic{table}}
\renewcommand\thefigure{S\arabic{figure}}
\renewcommand{\theHtable}{Supplement.\thetable}
\renewcommand{\theHfigure}{Supplement.\thefigure}

\subsection{Attribute-based Face Verification}

We present attribute-based face verification scores including False Non-Match Rate (FNMR), False match rate (FMR) and F1 score in the Table \ref{tab:tnr}. We use the same pairings and protocol \cite{LFWTech} presented in the Section \ref{sec:fver} for Table \ref{tab:fr1}. 

Whilst F1 scores are correlated with Table \ref{tab:fr1} accuracies, for the imbalanced training setup 1, the false matching ratio is higher on attributes like Monolid Eye, Type 6/5/4/3, Wide Nose, Full Lips than the different categories under the same attribute. Moreover, we observe that the balanced training setup 2 improves the FMR while increasing the FNMR for the attribute categories with higher accuracies and F1 scores.


\begin{table}[ht]
\small
\resizebox{\columnwidth}{!}{
\begin{tabular}{llll|lll}
\toprule

{} & \multicolumn{3}{c}{\thead{Setup 1}} & \multicolumn{3}{c}{\thead{Setup 2}} \\
\cmidrule(lr){2-7}
\thead{Attribute Name} &      \thead{F1} &     \thead{FNMR} &     \thead{FMR} &      \thead{F1} &     \thead{FNMR} &     \thead{FMR} \\
\midrule
Blonde Hair    &   96.85 &  1.40 &   3.83 &   96.04 &  2.53 &  3.13 \\
Red Hair       &   96.60 &  2.83 &   4.00 &   96.48 &  2.83 &  2.83 \\
Type 2         &   95.98 &  3.10 &   3.90 &   95.25 &  3.77 &  3.90 \\
Bald           &   95.32 &  5.00 &   5.37 &   95.44 &  3.23 &  4.97 \\
Gray Hair      &   95.00 &  3.70 &   5.87 &   95.93 &  2.53 &  5.13 \\
Brown Hair     &   94.46 &  6.40 &   4.63 &   94.08 &  5.43 &  4.43 \\
Type 6         &   94.42 &  4.10 &   7.30 &   95.01 &  5.53 &  4.67 \\
Wavy Hair      &   93.96 &  3.27 &   7.53 &   95.42 &  4.97 &  3.77 \\
Narrow Nose    &   93.05 &  6.77 &   7.20 &   94.29 &  4.40 &  5.87 \\
Type 5         &   92.72 &  4.07 &  10.33 &   94.45 &  5.63 &  5.47 \\
Curly Hair     &   92.51 &  5.47 &   9.67 &   93.58 &  6.87 &  5.70 \\
Small Lips     &   92.36 &  5.80 &   8.37 &   94.29 &  5.03 &  4.70 \\
Type 1         &   92.08 &  5.99 &   6.45 &   90.14 &  6.67 &  9.41 \\
Type 3         &   91.80 &  8.63 &   7.73 &   93.59 &  5.93 &  6.13 \\
Straight Hair  &   91.19 &  9.17 &   6.87 &   93.97 &  4.30 &  6.53 \\
Other Eye      &   91.16 &  7.23 &   7.27 &   93.76 &  7.43 &  4.47 \\
Wide Nose      &   90.99 &  7.23 &   7.43 &   89.78 &  7.10 &  5.27 \\
Full Lips       &   90.73 &  6.60 &  10.17 &   93.43 &  7.13 &  5.77 \\
Type 4         &   90.45 &  8.30 &   8.53 &   93.50 &  5.70 &  6.93 \\
Black Hair     &   90.12 &  7.77 &   8.73 &   90.50 &  6.83 &  5.83 \\
Monolid Eye    &   88.84 &  9.53 &  13.03 &   90.62 &  8.47 &  6.93 \\
\bottomrule

\end{tabular}
}
\vspace{0.1cm}
\caption{Attribute-based face verification F1, FNMR, TMR scores of RFW dataset on both training setups.}
\label{tab:tnr}
\end{table}

\end{document}